\documentclass{article} 
\usepackage{iclr2026_delta,times}

\usepackage{amsmath,amsfonts,bm}

\def\eqref#1{equation~\ref{#1}}

\def\1{\bm{1}}

\def\eps{{\epsilon}}

\DeclareMathAlphabet{\mathsfit}{\encodingdefault}{\sfdefault}{m}{sl}
\SetMathAlphabet{\mathsfit}{bold}{\encodingdefault}{\sfdefault}{bx}{n}

\usepackage{hyperref}
\usepackage{graphicx}
\usepackage{url}
\iclrfinalcopy

\title{Curriculum Sampling: A Two-Phase Curriculum for Efficient Training of Flow Matching}

\author{Pengwei Sun  \\
Stanford University\\
\texttt{pengwei@stanford.edu} \\
}

\begin{document}

\maketitle

\begin{abstract}
Timestep sampling $p(t)$ is a central design choice in Flow Matching models, yet common practice increasingly favors static middle-biased distributions (e.g., Logit-Normal). We show that this choice induces a speed--quality trade-off: middle-biased sampling accelerates early convergence but yields worse asymptotic fidelity than Uniform sampling. By analyzing per-timestep training losses, we identify a U-shaped difficulty profile with persistent errors near the boundary regimes, implying that under-sampling the endpoints leaves fine details unresolved. Guided by this insight, we propose \textbf{Curriculum Sampling}, a two-phase schedule that begins with middle-biased sampling for rapid structure learning and then switches to Uniform sampling for boundary refinement. On CIFAR-10, Curriculum Sampling improves the best FID from $3.85$ (Uniform) to $3.22$ while reaching peak performance at $100$k rather than $150$k training steps. Our results highlight that timestep sampling should be treated as an evolving curriculum rather than a fixed hyperparameter.
\end{abstract}

\section{Introduction}
Flow-based generative models, such as Flow Matching (FM), have fundamentally shifted the landscape of continuous-time generative modeling \citep{lipman2022flow, liu2022flow}. By defining a probability path that transforms noise into data via a velocity field, these models offer a stable, simulation-free training objective that scales effectively to high-resolution image generation \citep{esser2024scaling}. 

In Flow Matching, a neural network is trained to approximate a time-dependent velocity field by sampling a time step $t \sim p(t)$ and regressing an appropriate training target defined by the chosen probability path.
At the same time, the design space around this training objective is rapidly expanding with emerging techniques. For example, Mean Flow \citep{meanflow} replaces instantaneous velocities with \emph{average} velocities to enable strong one-step generation, improved Mean Flow variants \citep{imf} reformulate the objective and guidance mechanism for better stability and flexibility, and ``Just image Transformers'' \citep{jit} advocate predicting clean images $x$ directly (instead of noise/noised quantities) using large-patch Transformers. These approaches use different loss functions, prediction parameterizations (denoising vs. $x$-prediction), or different types of velocity (instantaneous vs. average).
Nevertheless, they all share a common requirement: training must sample timesteps from some distribution $t \sim p(t)$ on $[0,1]$. A fundamental component shared across this entire family of models is therefore the choice of the time step sampling distribution $p(t)$ used during training.

The importance of the time step sampling strategy is universal. Regardless of whether the objective is to train a standard model for multi-step inference (multi-NFE, NFE stands for number of function evaluations) or a specialized model optimized for extremely fast generation (1-NFE or 2-NFE), the model must learn to approximate the vector field across the entire interval $t \in [0, 1]$. Similarly, this requirement holds true whether the model targets the instantaneous velocity field or the average velocity field. The choice of $p(t)$ dictates the allocation of computational budget across the trajectory, implicitly weighting the model's focus between the chaotic "transport" phase in the middle and the boundary conditions at the noise ($t=1$) and data ($t=0$) limits.

In this study, we revisit the standard assumptions regarding these sampling distributions. Through extensive experimentation, we identify a trade-off between convergence speed and final generation quality in existing timestep sampling strategies. We show that while Logit-Normal sampling accelerates early training, it imposes a performance ceiling due to under-sampling of the trajectory boundaries. We provide a detailed analysis of the training dynamics, characterizing a U-shaped loss landscape where the difficulty of velocity prediction peaks at the data ($t \to 0$) and noise ($t \to 1$) extremes, necessitating explicit attention to these regimes for high-fidelity results. We introduce curriculum sampling, a two-phase curriculum that transitions from a structure-focused distribution (Logit-Normal) to a coverage-focused distribution (Uniform). This method outperforms static baselines, achieving a 16\% relative improvement in FID over standard Uniform sampling and a 33\% convergence speed-up.

\section{Preliminary: Flow Matching}
Flow Matching (FM) \citep{lipman2022flow,liu2022flow,albergo2022building} is a family of generative models that learn a time-dependent velocity field transporting a simple prior distribution into the data distribution. During training, the model samples a time step $t \sim p(t)$ on $[0,1]$; the choice of $p(t)$ determines which parts of the trajectory are emphasized. Formally, given data $x \sim p_\text{data}(x)$ and prior noise $\eps \sim p_\text{prior}(\eps)$, one constructs a path $z_t$ that interpolates between $x$ and $\eps$ over time $t \in [0,1]$:
\begin{equation}
    z_t = a_t x + b_t \eps,
\end{equation}
where $a_t$ and $b_t$ are predefined schedules. Differentiating w.r.t. time gives the corresponding \emph{conditional} velocity
\begin{equation}
    v_t = \frac{d}{dt} z_t = a'_t x + b'_t \eps,
\end{equation}
which is denoted by $v_t = v_t(z_t \mid x)$ in \citet{lipman2022flow}. A commonly used schedule is $a_t = 1 - t$ and $b_t = t$, yielding $v_t = \eps - x$.

Because the same point $z_t$ may arise from multiple pairs $(x,\eps)$, FM is concerned with the \emph{marginal} velocity field, defined as the conditional expectation over all possible pairings:
\begin{equation}
v(z_t, t) \stackrel{\text{def}}{=} \mathbb{E}_{p_t(v_t\mid z_t)}\big[ v_t \big].
\label{eq:instant_v}
\end{equation}
A neural network $v_\theta$ parameterized by $\theta$ is trained to fit this marginal field via
\begin{equation}
\mathcal{L}_\text{FM}(\theta) = \mathbb{E}_{t,\,z_t \sim p_t(z_t)} \big\| v_\theta(z_t, t) - v(z_t, t) \big\|^2.
\end{equation}
Directly evaluating $\mathcal{L}_\text{FM}$ is typically infeasible because \eqref{eq:instant_v} requires marginalization. \citet{lipman2022flow} propose to instead optimize the \emph{conditional} Flow Matching objective (often referred to as Conditional Flow Matching, CFM):
\begin{equation}
\mathcal{L}_\text{CFM}(\theta) = \mathbb{E}_{t,\,x,\,\eps} \big\| v_\theta(z_t, t) - v_t(z_t \mid x) \big\|^2,
\end{equation}
and show that minimizing $\mathcal{L}_\text{CFM}$ is equivalent to minimizing $\mathcal{L}_\text{FM}$.

Given a learned marginal velocity field $v(z_t, t)$, sampling is performed by solving the ODE
\begin{equation}
\frac{d}{dt} z_t = v(z_t, t),
\label{eq:fm_ode}
\end{equation}
starting from $z_1 = \eps \sim p_\text{prior}$. In practice, the integral solution is approximated numerically using a discrete-time solver (e.g., Euler or higher-order methods).

\section{Related Work}
\textbf{Timestep Sampling Strategies.}
The choice of the sampling distribution $p(t)$ is crucial for the efficiency of flow-based models. While Uniform sampling is the standard baseline \citep{Liu2022FlowSA, Liu2023InstaFlowOS, lipman2022flow}, recent work has explored non-uniform strategies that reallocate compute across the trajectory.
 Logit-Normal sampling, which concentrates density in the middle of the interval ($t \approx 0.5$), is widely adopted in rectified flow models \citep{esser2024scaling} and other recent flow-based models \citep{meanflow, jit, imf}.

 \citet{lee2024improvingtrainingrectifiedflows} argue that the best choice of $p(t)$ is not universal: it depends on what the model is asked to predict near the endpoints. For 1-rectified flow training, the endpoints can be comparatively uninformative (e.g., predicting dataset/noise averages), so the learning signal is often concentrated in the middle of the interval. In contrast, for higher-order rectified flows (e.g., 2-rectified flow), the endpoint objectives become nontrivial, and empirical analyses report a boundary-heavy (U-shaped) difficulty profile. This motivates timestep distributions that deliberately oversample $t$ near $0$ and $1$ rather than focusing solely on the transport phase.

\section{Methodology}
\subsection{Proposed Method: Curriculum Sampling}

Based on the empirical evidence, we propose a two-phase curriculum strategy for timestep sampling during training:

\begin{enumerate}
    \item \textbf{Structure Phase:} Use Logit-Normal sampling with inverted U-shape distribution. This distribution concentrates training samples in the middle regime (around t=0.5), while maintaining enough variance to cover the transport phase. This maximizes the rate of FID reduction and ensures sufficient training at mid-$t$.

    \item \textbf{Refinement Phase:} Switch to Uniform sampling. Once the global structure is established, the model must reduce residual errors that persist near the boundary regimes; Uniform sampling ensures these regions are revisited with sufficient frequency.

\end{enumerate}

We choose three families of sampling strategies:
\begin{enumerate}
    \item \textbf{Uniform Baseline:} $t \sim \mathcal{U}(0, 1)$.
    \item \textbf{Mode Sampling:} A parametric family favoring specific modes, defined by a symmetric Beta distribution:
    \begin{equation}
        p(t) \propto t^s (1-t)^s, \quad s > -1
    \end{equation}
    We evaluated $s \in \{-0.5, 1.0, 1.2\}$. Notably, $s=-0.5$ corresponds to the Arcsine distribution ($Beta(0.5, 0.5)$), creating a U-shaped density that biases sampling toward the boundaries. Positive $s$ values concentrate density in the middle ($t=0.5$) while down-weighting the endpoints.
    \item \textbf{Logit-Normal (LN) Sampling:} Timesteps are sampled by applying a sigmoid transformation to a normal variable \citep{logitnormal}:
    \begin{equation}
        t = \text{sigmoid}(u) = \frac{1}{1 + e^{-u}}, \quad u \sim \mathcal{N}(\mu, \sigma^2)
    \end{equation}
    The corresponding probability density function is:
    \begin{equation}
        p(t) = \frac{1}{\sigma t (1-t) \sqrt{2\pi}} \exp\left( -\frac{(\text{logit}(t) - \mu)^2}{2\sigma^2} \right)
    \end{equation}
    where $\text{logit}(t) = \ln\left(\frac{t}{1-t}\right)$. We tested several variations including $\mu \in \{-1.0, -0.8, -0.6, -0.4, -0.1, 0.8\}$ and $\sigma \in \{0.5, 1.0\}$ to control the center and spread of the distribution. LN sampling allocates very little density near the boundaries, which differs from the Mode sampling ($s>0$).
\end{enumerate}

\subsection{Experimental Setup}
We trained a series of Conditional Flow Matching (CFM) models for class-unconditional image generation on CIFAR-10. The input to the model is $32\times32\times3$ in pixel space. All models shared the same architecture and optimization hyperparameters, varying only the timestep sampling distribution $p(t)$. We use the
U-Net \citep{10.1007/978-3-319-24574-4_28} architecture developed from \citep{NEURIPS2019_3001ef25} (about 55M parameters), as it is commonly used in prior work \citep{meanflow, salimans2022progressivedistillationfastsampling}.  
We use batch size 1024, Adam \citep{kingma2017adammethodstochasticoptimization} with learning rate 0.0006, $(\beta_1, \beta_2) = (0.9, 0.999)$, weight decay 0, and EMA decay of 0.99995. The model is trained for 150K iterations with 10K warm-up \citep{goyal2018accuratelargeminibatchsgd}. The power $p$ for adaptive weighting is 0.75.
During inference, 50-NFE Euler solver is used to solve the ODE.
Performance was measured using Fréchet Inception Distance (FID) computed on 50k generated samples.

\section{Results}

\subsection{Baseline Performance Comparison}
\subsubsection{Best FID of baseline strategies}
\begin{figure}[t]
    \centering
    \includegraphics[width=0.8\linewidth]{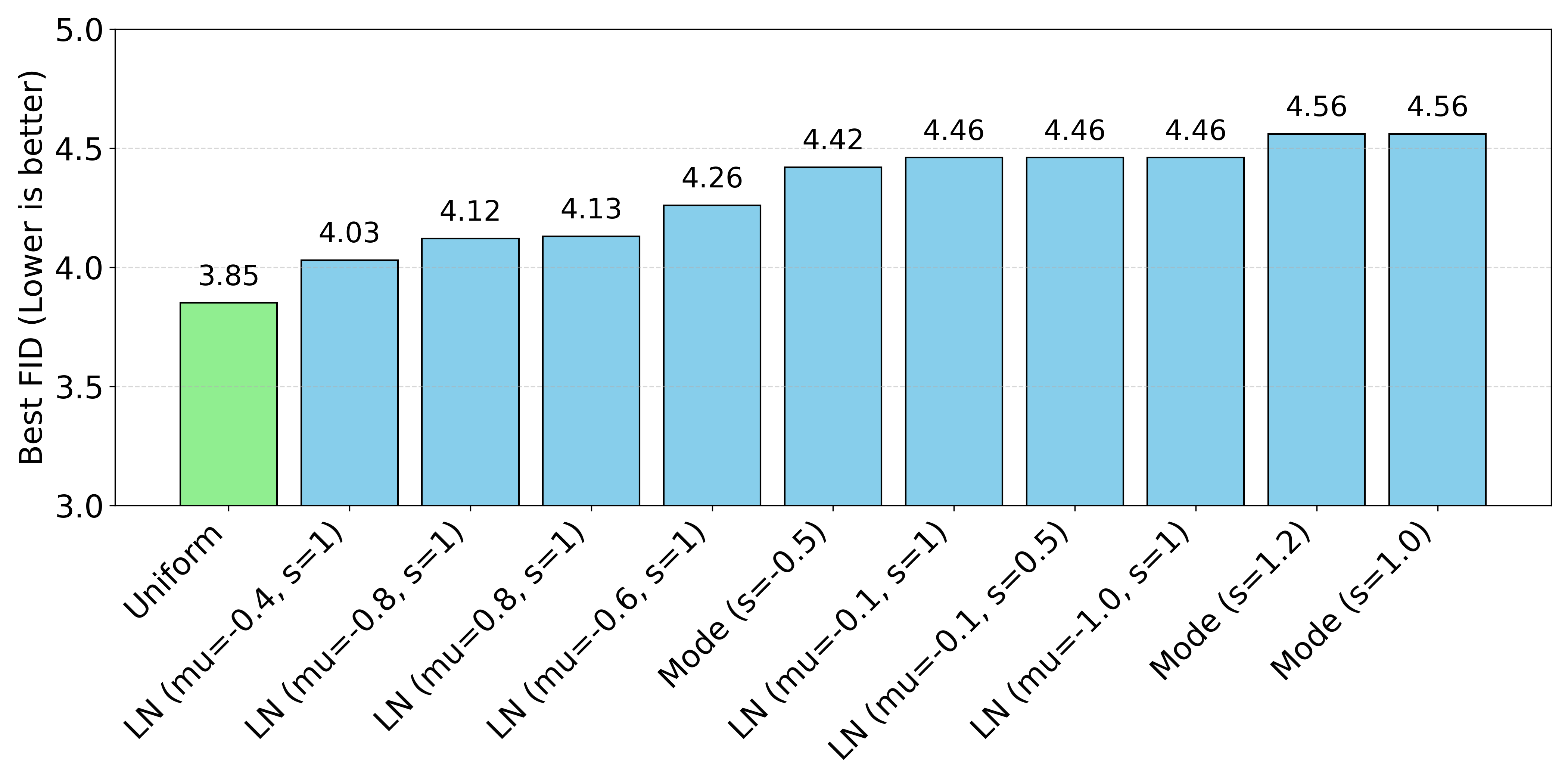}
    \caption{Comparison of best achieved FID scores across non-curriculum sampling strategies. Methods are sorted from best (lowest FID) to worst.}
    \label{fig:best_fid_comparison}
\end{figure}
We begin by establishing a performance baseline across canonical time-step sampling strategies without curriculum learning. Figure \ref{fig:best_fid_comparison} presents the minimum Fréchet Inception Distance (FID) achieved by each method throughout the training process.
The results demonstrate that simple sampling strategies remain highly competitive. The Uniform baseline achieved the lowest FID of 3.85, outperforming all biased strategies. Among the non-uniform methods, Logit-Normal distributions modestly skewed toward the data (image) end performed best: LN with $\mu=-0.4$ reached an FID of 4.03, while $\mu=-0.8$ reached 4.12. Narrower or more centrally focused distributions generally yielded higher (worse) FID scores.
Mode sampling with a U-shaped density ($s=-0.5$) yields the worst FID among the tested settings.
Overall, the results suggest that while emphasizing the transport regim (mid-$t$) performs better than focusing on the boundary timesteps, broad coverage of the entire time domain is essential for maximizing final generation fidelity.

\subsubsection{Convergence Dynamics of baseline strategies}
\begin{figure}[t]
    \centering
    \includegraphics[width=0.65\linewidth]{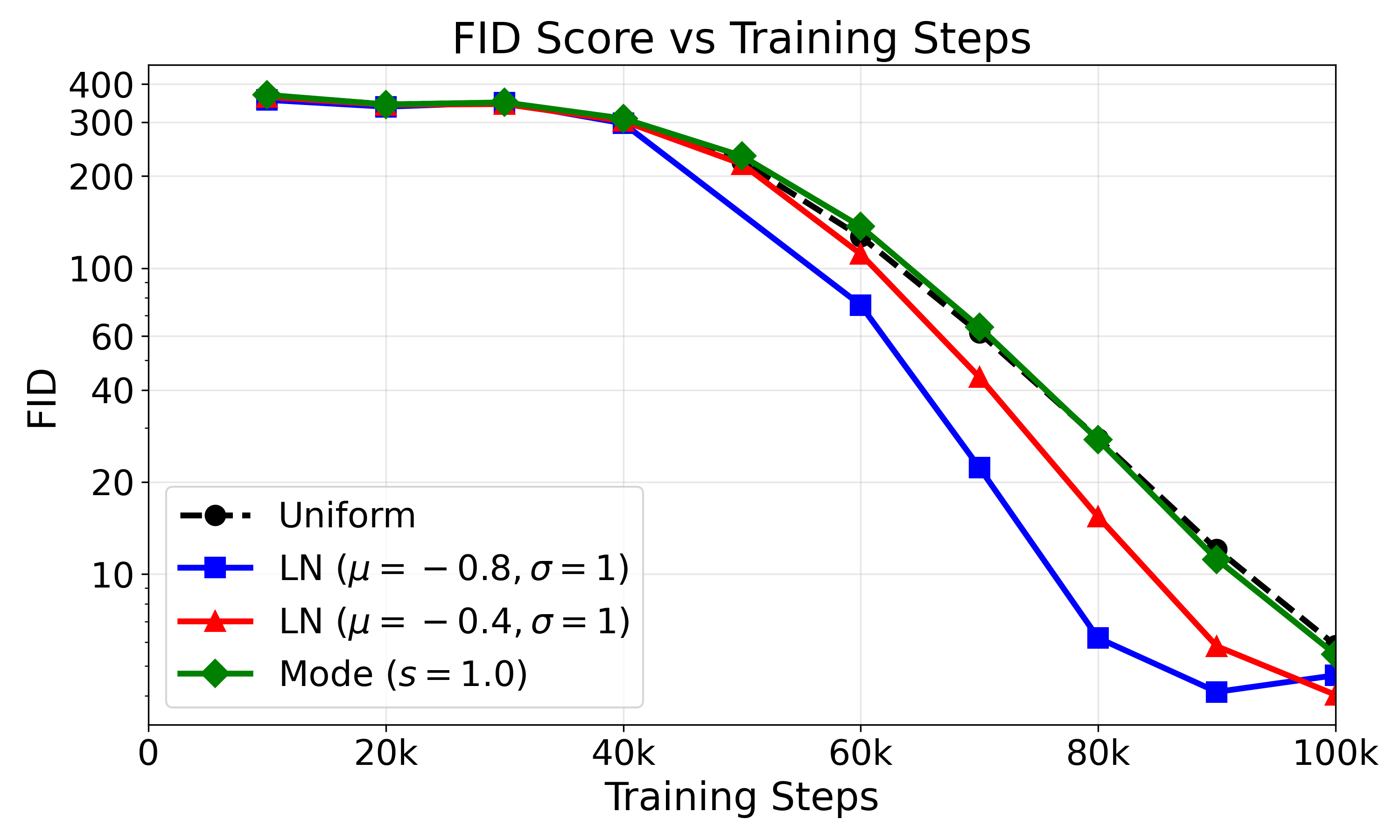}
    \caption{Comparison of Fréchet Inception Distance (FID) scores over the first 100k training steps for Uniform sampling versus Logit-Normal (LN, $\mu \in \{-0.8, -0.4\}, \sigma=1$) and Mode ($s=1.0$) sampling.}
    \label{fig:fid_trajectories}
\end{figure}

Figure \ref{fig:fid_trajectories} visualizes the FID convergence for Uniform, Mode, and Logit-Normal sampling over the first 100k training iterations,
revealing a distinct trade-off between convergence speed and asymptotic fidelity. We observe that
inverted U-shape distributions, such as Logit-Normal ($\mu=-0.4$ or $-0.8$), act as effective accelerators, yielding significantly lower FID scores in the early training regime from 40k to 90k steps
compared to the Uniform baseline.
This supports the hypothesis that reallocating compute toward mid-$t$ regime can facilitate rapid structural learning.
However, this advantage is transient; as training matures,
the improvement rate of biased methods saturates, leading to worse final FID (Fig. \ref{fig:best_fid_comparison}), and sometimes overfitting (Fig. \ref{fig:fid_trajectories}). In contrast, the Uniform
baseline maintains a steady rate of improvement, eventually surpassing the biased strategies. This crossover
phenomenon highlights that while ''middle-bias'' accelerates the learning of the vector field's global mode, it
systematically down-samples the boundary regimes ($t \to 0, 1$), thereby imposing a ceiling on the model's final
generation quality.
\subsubsection{Loss Landscape Analysis}
\label{sec:loss_landscape}
We monitor the time-dependent training loss $\mathcal{L}(\tilde{t})$ throughout the training process, which is defined as:
\begin{equation}
\mathcal{L}(\tilde{t}) = \mathbb{E}_{x,\eps} \big\| v_\theta(z_{\tilde{t}}, \tilde{t}) - v_{\tilde{t}}(z_{\tilde{t}} \mid x) \big\|^2,
\end{equation}

\begin{figure*}[t]
\centering
\includegraphics[width=\linewidth]{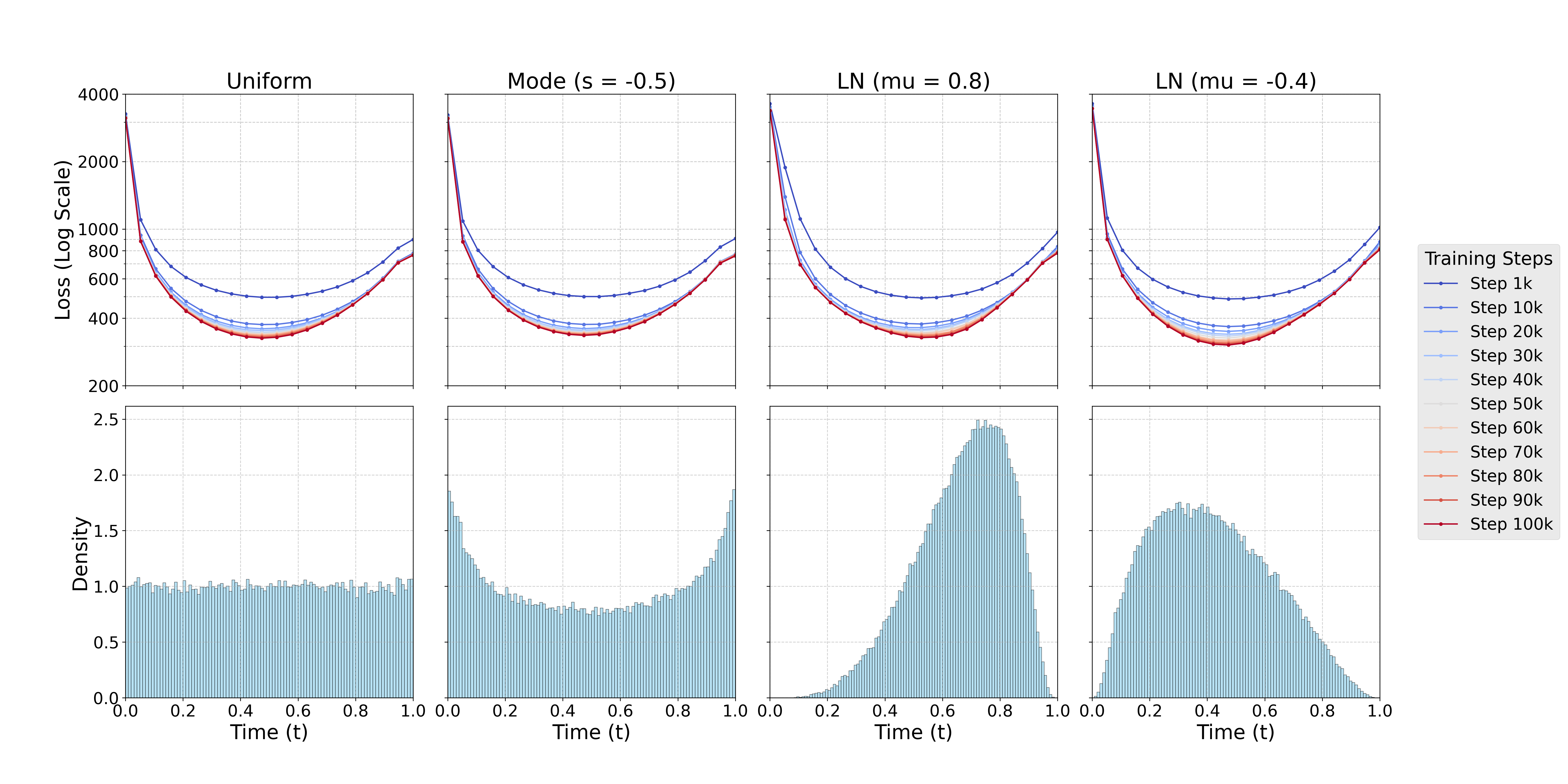}
\caption{Time-Dependent Training Loss Dynamics and Sampling Densities.
Top row: Evolution of the loss over the diffusion time $t \in [0, 1]$ throughout training (color gradient from 10k to 100k steps).
Bottom row: Histograms of sampled time steps $t \sim p(t)$ from each distribution.
Columns represent: (a) Uniform sampling, (b) Mode sampling ($s=-0.5$), which skews towards $t=0$ and $t=1$, (c) Logit-Normal sampling ($\mu=0.8$), shifting focus towards the noise-dominant regime ($t$ close to 1), and (d) Logit-Normal sampling ($\mu=-0.4$), emphasizing the mid-trajectory.
}
\label{fig:loss_dynamics}
\end{figure*}

Figure~\ref{fig:loss_dynamics} visualizes the evolution of the training loss and the corresponding sampling density across different baselines. Consistent with findings in \cite{lee2024improvingtrainingrectifiedflows}, the training loss exhibits a distinctive U-shape with respect to the time step $t$, reflecting the inherent difficulty of velocity prediction at different stages of the trajectory.
The loss is largest near the data end ($t \approx 0$) and remains elevated toward the noise end ($t \approx 1$). This suggests that learning to form the final sharp details of the image is the most challenging part of the objective, followed by the initial departure from the noise. The middle transport phase ($t \approx 0.5$) has the lowest loss regardless of the timestep sampling, implying the vector field there is smoother or easier to fit.
Loss reduction is also slower at the boundaries ($t \to 0$ and $t \to 1$), suggesting that accurately predicting the velocity field becomes progressively harder near the data (images) and noise endpoints.

Logit-Normal sampling ($\mu=-0.4$), by concentrating probability mass in the middle of the trajectory, accelerates loss reduction in this easier regime but at the cost of slower convergence at the boundaries, potentially leading to insufficient training in these critical regions. Conversely, U-shaped Mode sampling ($s=-0.5$) over-emphasizes the boundaries while neglecting the middle trajectory, which explains its suboptimal performance as it fails to adequately learn the core transport mapping. These observations underscore the necessity of a balanced sampling strategy or a curriculum that can adaptively shift focus from learning the global structure (middle $t$) to refining the boundary conditions.

\subsection{Optimizing the Curriculum}
Motivated by the distinct strengths of different sampling strategies, we systematically optimized a two-phase curriculum. This curriculum is defined by a switching time $T_s$, a first-phase distribution $p_1(t)$, and a second-phase distribution $p_2(t)$.

\subsubsection{Curriculum Learning Outperforms Static Baselines}
  Our empirical results demonstrate that dynamic time step sampling strategies consistently surpass static baselines (Table \ref{tab:fid_phases}). 
  The proposed curriculum approach achieves the best FID of 3.22, representing a substantial 16.4\%
  relative improvement over the standard Uniform baseline (FID 3.85) and a significant gain over the static
    Logit-Normal baseline (FID 4.12). This supports the hypothesis that different timestep regimes benefit from different amounts of training; the optimal sampling distribution can be non-stationary and should evolve during training.

    The Uniform baseline achieves its best FID at 150k steps. In contrast, curricula starting with LN ($\mu=0.8$) peak much earlier at 100k. This indicates that the curriculum strategy primarily acts as an accelerator, allowing the model to reach a higher quality solution with significantly fewer training steps (33\% fewer).
  
\subsubsection{Phase 1 Distribution ($p_1(t)$)}
We evaluate three Logit-Normal (LN) configurations for the phase-1 timestep distribution. These choices accelerate early training relative to Uniform sampling (Fig.~\ref{fig:fid_trajectories}), but differ in where they place probability mass. In the curriculum setting, LN($\mu=0.8$) performs best, followed by LN ($\mu=-0.4$) and LN ($\mu=-0.8$). LN ($\mu=0.8$) places more mass near $t=1$, whereas the other two emphasize timesteps closer to $t=0$. This suggests that, for our setup, allocating more updates toward the noise-side regime in phase 1 can lead to faster convergence and better downstream performance.
\subsubsection{Phase 2 Distribution (\texorpdfstring{$p_2(t)$}{p2(t)})}
For the refinement phase, we consider Uniform sampling and Mode sampling ($s=-0.5$, the U-shaped ``bathtub'' distribution) to increase sampling at boundary timesteps. A critical finding is that curricula that transition into either a Uniform or a Mode distribution in phase-2 outperform the corresponding static distributions.

Comparing LN($\mu=-0.8$) $\to$ Uniform (FID 3.67) against the static LN($\mu=-0.8$) baseline (FID 4.12) shows that re-introducing sampling density at the boundaries ($t \to 0$ and $t \to 1$) is essential in the late stage of training. Comparing the top-performing LN $\to$ Uniform strategy (FID 3.22) against the LN $\to$ Mode strategy (FID 3.42) further indicates that while boundary refinement is critical, over-sampling the boundaries at the expense of the transport path is less effective.

\subsubsection{Switching Time (\texorpdfstring{$T_s$}{Ts})}
We find that the curriculum switch time is a sensitive hyperparameter. Across runs, the best performance is consistently achieved at $T_s \approx 60$k iterations (roughly 40\% of the training budget). Switching too early or too late yields suboptimal convergence, reflecting a trade-off between the structural acceleration phase and the boundary refinement phase. For example, early switches (e.g., 50k) do not fully leverage the convergence acceleration from Logit-Normal sampling and leave insufficient training in the mid-$t$ regime, whereas late switches (e.g., 70k) allocate insufficient time for the boundary refinement needed to lower the FID floor.

\begin{table}[t]
\caption{Comparison of Best FID Scores with different curriculum settings. The results are ranked from low to high (lower is better). The last two entries are static sampling without curriculum.}
\label{tab:fid_phases}
\begin{center}
\begin{tabular}{lllcc}
\multicolumn{1}{c}{$p_1(t)$} & \multicolumn{1}{c}{\bf $p_2(t)$} & \multicolumn{1}{c}{\bf $T_s$} & \multicolumn{1}{c}{\bf Training Step @ Best FID} & \multicolumn{1}{c}{\bf Best FID}
\\ \hline \\
LN ($\mu=0.8$) & Uniform & 60k & 100k & \textbf{3.22}\\
LN ($\mu=-0.4$) & Uniform & 60k & 100k & 3.24 \\
LN ($\mu=0.8$) & Uniform & 50k & 100k & 3.34 \\
LN ($\mu=0.8$) & Mode & 60k & 100k & 3.42 \\
LN ($\mu=0.8$) & Mode & 90k & 100k & 3.43 \\
LN ($\mu=0.8$) & Uniform & 70k & 100k & 3.45 \\
LN ($\mu=0.8$) & Uniform & 80k & 100k & 3.56 \\
LN ($\mu=-0.8$) & Uniform & 60k & 150k & 3.67 \\
LN ($\mu=-0.8$) & Uniform & 80k & 150k & 3.72 \\
Uniform & Uniform & - & 150k & 3.85 \\
LN ($\mu=-0.8$) & LN ($\mu=-0.8$) & - & 90k & 4.12 \\
\end{tabular}
\end{center}
\end{table}

\section{Conclusion}
This work revisits a central but often under-specified design choice in flow-based generative modeling: the timestep sampling distribution $p(t)$. We identify a performance paradox in standard training: static, middle-biased sampling (e.g., Logit-Normal) can substantially accelerate early convergence, yet it can underperform Uniform sampling in final sample quality because it systematically under-samples the boundary regimes $t\to 0$ and $t\to 1$. Empirically, we confirm this mechanism by analyzing the time-dependent loss landscape, which exhibits a pronounced U-shape with the largest residual errors concentrated near the data and noise endpoints.

Motivated by this analysis, we propose a two-phase curriculum for timestep sampling. The curriculum uses a Logit-Normal distribution in a structure-learning phase to exploit fast early progress, then switches to Uniform sampling in a refinement phase to restore coverage of the boundaries. On CIFAR-10, this simple schedule improves the best FID from $3.85$ (Uniform) to $3.22$ (a $16.4\%$ relative improvement) while reaching its best performance at $100$k steps instead of $150$k (a $33\%$ reduction in training steps). Beyond this specific instantiation, our results emphasize that timestep sampling is not a one-shot hyperparameter: matching $p(t)$ to the evolving difficulty across time can be as consequential as architectural choices, and suggests a broader opportunity for dynamic and adaptive training protocols in flow-based and related continuous-time generative models.

Finally, our findings may also inform inference-time timestep allocation. Since approximation error is not uniform across $t$, a potential direction is to design inference schedules (e.g., non-uniform discretizations or adaptive step placement) that devote more resolution to the time regimes where the learned vector field is hardest to approximate.

\bibliography{iclr2026_delta}
\bibliographystyle{iclr2026_delta}


\end{document}